\pdfoutput=1
\documentclass{ifacconf}

\usepackage{graphicx}      
\usepackage{natbib}        
\usepackage{amsmath}
\usepackage{amssymb}
\usepackage{booktabs}
\usepackage{tabularx}
\usepackage{tikz}
\usepackage{url}

\usepackage{eso-pic}

\AddToShipoutPictureBG*{%
  \AtPageUpperLeft{%
    \setlength\unitlength{1mm}%
    \put(0,-15){%
      \makebox(\paperwidth,0)[c]{%
        \fbox{\footnotesize\textbf{This work has been submitted to IFAC for possible publication.}}%
      }%
    }%
  }%
}

\begin{document}
\begin{frontmatter}

\title{Behavioral Evaluation of Large Language Models Under Cognitive Biases in HAZOP Studies \thanksref{footnoteinfo}}  

\thanks[footnoteinfo]{This work was supported by \textit{I.Systems Automação Industrial S.A} through the financial resources provided to conduct the experiments.}

\author[feq]{Vinicius M. Müller} 
\author[feq]{Vitor A. O. Silva}
\author[feq]{Sávio S. V. Vianna}
\author[feq]{Flávio V. Silva}

\address[feq]{School of Chemical Engineering, University of Campinas (UNICAMP), Campinas, SP, Brazil, (email: flaviovs@unicamp.br)}



\begin{abstract}                
Hazard and Operability (HAZOP) studies safeguard industrial processes against catastrophic accidents. Large language models (LLMs) increasingly assist HAZOP drafting, yet evaluations rely on lexical overlap. Because LLMs internalize human heuristics, prompts may trigger biased judgments. Whether frontier LLMs withstand cognitive perturbations in safety-critical workflows remains unknown. Here we show across 420 trials that three frontier LLMs resisted $96.8\%$ of cognitive manipulations, rendering overall susceptibility statistically inconclusive. Contrary to assumed broad fragility, models rejected confirmation and availability cues. However, safety-critical risk downgrades emerged under low anchors and gain frames. Thus, automated hazard analysis requires disciplined human-in-the-loop validation.
\end{abstract}


\end{frontmatter}

\section{Introduction}


Hazard and Operability (HAZOP) analysis remains the cornerstone qualitative methodology for process hazard analysis (PHA) across the process industry, serving as an indispensable systematic safeguard against catastrophic accidents and operational deviations \citep{dunjo_hazard_2010, cameron_process_2017}. Traditionally executed through multidisciplinary expert workshops, the conventional HAZOP procedure requires a thorough, line-by-line examination of Piping and Instrumentation Diagrams (P\&IDs) using standardized guidewords.

However, this manual approach imposes substantial temporal overhead, steep economic costs, and severe cognitive strain on participating specialists \citep{cameron_process_2017}. The repetitive and exhaustive nature of these sessions often induces cognitive fatigue, exposing the analysis to heuristic biases, missed failure modes, and potential scenario omissions \citep{baybutt_cognitive_2016}. To mitigate these procedural vulnerabilities, researchers have pursued automated and computer-aided HAZOP frameworks for over three decades, progressively exploring rule-based expert systems, qualitative fault-propagation digraphs, and formal domain ontologies \citep{cameron_process_2017, elhosary_automation_2024}. Although these early knowledge-engineered systems demonstrated capability in tracing causal dependencies, their practical scalability remained constrained by the brittle curation of static rule bases and an inability to parse unstructured engineering narratives \citep{elhosary_automation_2024}.

The emergence of Transformer architectures and Large Language Models (LLMs) has catalyzed a fundamental paradigm shift from deterministic inference engines to generative, data-driven synthesis. Foundation models demonstrate unprecedented facility in contextual reasoning across technical documentation, offering viable mechanisms to assist or semi-automate the drafting of process safety worksheets directly from process design specifications \citep{elhosary_automation_2024, lee_can_2026}.

Notwithstanding these preliminary advances, current evaluations of generative models in process hazard identification remain predominantly anchored to lexical overlap and surface-level embedding similarities (e.g., ROUGE, BERTScore), a methodology that fails to assess the logical consistency and underlying behavioral robustness of model-generated safety conclusions \citep{lee_can_2026, elhosary_automation_2024}. This evaluation paradigm critically overlooks a foundational reality long recognized in behavioral safety engineering: human expert judgment within PHA and HAZOP sessions is inherently prone to systematic cognitive shortcuts and inferential heuristics \citep{baybutt_cognitive_2016, baybutt_validity_2018}.

Empirical observations show that safety review teams regularly succumb to anchoring effects established by authoritative peers, confirmation bias that defaults to assumptions of nominal operator compliance, and availability heuristics that overweigh recent salient incidents while neglecting complex multivariable interactions \citep{baybutt_cognitive_2016, baybutt_validity_2018}. Because foundation LLMs are pretrained on extensive human corpora and aligned through reinforcement learning from human feedback (RLHF), they do not function as neutral, objective safety arbiters. Rather, they absorb, replicate, and often amplify these anthropomorphic inferential traps \citep{cheung_large_2025}. When prompted, generative models exhibit pronounced polarity asymmetries, sycophantic agreement, and framing sensitivities capable of inverting risk classifications under identical physical boundary conditions based solely on subtle rhetorical cues \citep{cheung_large_2025}. Despite the severe consequences of flawed risk characterization in high-hazard facilities, the literature lacks empirical investigations that systematically stress-test the behavioral resilience of frontier language models against standardized cognitive perturbations within safety-critical workflows.

To address this gap, this study presents a systematic behavioral characterization of frontier LLMs subjected to controlled cognitive perturbations in process hazard identification tasks. Adopting a counterbalanced between-subjects experimental paradigm, we evaluate model resilience across standardized HAZOP cases derived from an industrial reference topology, directly isolating prompt-level contextual framing. To eliminate conversational confounders and ensure reproducible baselines, inference is executed under stateless, single-pass queries at temperature zero (T = 0). Technical outputs, encompassing deviation causes, consequence severities, and safeguard allocation, are benchmarked through double-blind expert scoring against established process safety standards. To account for intrinsic baseline heterogeneity across the node, we formulate a Bayesian generalized linear mixed model (GLMM) with case-specific random intercepts \citep{Fong2010}, estimating posterior distributions of conditional manipulation effects on the log-odds of a biased technical judgment. By decoupling prompt-driven heuristics from intrinsic domain capabilities, this framework provides a rigorous statistical appraisal of generative model stability in safety-critical operations.

We formulate two preregistered hypotheses concerning bias susceptibility and model-level heterogeneity:

\begin{hypo}[Bias Susceptibility] \label{hip:1}
Under \emph{quasi} deterministic\footnote{Even with greedy decoding (temperature $T=0$), LLM outputs are not strictly deterministic due to low-level hardware and infrastructural factors \citep{Atil2024}.} single-pass inference on standardized HAZOP tasks, prompt-level cognitive cues increase the conditional log-odds of unsupported biased judgments relative to neutral prompts.
\end{hypo}

\begin{hypo}[Model-Specific Vulnerability] \label{hip:2}
    \ The magnitude of behavioral susceptibility to prompt-level cognitive cues differs systematically across evaluated language models.
\end{hypo}

\section{Methodology}

\subsection{Reference System}
The controlled computational experiment used Node~1 of the crude-oil production HAZOP reported by \citet{Penelas2021}: the Free Water Knock-Out (FWKO) separator V-001 and associated lines. Tables~A3--A9 of \citet{Penelas2021} informed case construction and expert evaluation. Although \citet{Penelas2021} analyzed six nodes, complete HAZOP worksheets detailing causes, consequences, safeguards, and risk classifications were published exclusively for Node~1. The source risk matrix was treated as qualitative and ordinal, not as a numerical probability scale. Of 13 deviation records in Node~1, ten supported standalone cases; three lacked separately characterized scenarios (reverse flow, higher temperature, and fire/explosion) and received no invented risk ratings. Source-derived engineering facts were distinguished from synthetic vignettes, diagnostic observations, intervention assumptions, and outcome realizations.

The purposively constructed battery comprised 47 cases, 15 condition cells, and 140 response events in 56 contrast groups per model. Anchoring (10 cases $\times$ 3 conditions: control, low, high anchor; 30 events) and confirmation (10 $\times$ 2: control, operator-blame hypothesis; 20 events) covered all ten applicable deviations. Omissions in other families were structural rather than post-hoc: hindsight (9 $\times$ 2: foresight, hindsight; 18 events) omitted higher flow because the source table documented no consequences or safeguards to ground an adverse outcome; framing (9 $\times$ 3 conditions: neutral, gain, loss, each crossed with approve/reject wording; 54 events) omitted higher level because the source listed no safeguard recommendation to evaluate; and availability (9 $\times$ 2: control, salient vignette; 18 events) paired nine vignettes with a complete-node review, retaining the three non-applicable entries for explicit recognition while using the documented FWKO start/stop procedural item as an unprimed benchmark. Because the complete-node review does not depend on the vignette's deviation, the nine availability control prompts differed only in event identifier and were not independent baselines. They were nevertheless retained so that every vignette formed a complete contrast group with a control from its own case, preserving the uniform group-wise design, rating, and complete-group inclusion rule across families; the effective number of independent availability baselines is therefore one, which limits the precision of availability contrasts. The battery was neither a probability sample of installations nor sized through a formal power calculation. Prompt templates, response schemas, and the rating codebook are available at \url{https://github.com/ViniMelloeMuller/psyhazop-supplementary-material} v0.1.

The design assigned one generation request to each case--condition--question-form combination per model. A contrast group contained all conditions for one case, model, and question form. Within groups, engineering facts and response schemas were held constant, while manipulations and associated instructions varied. Contrasts therefore concerned complete prompt differences, including the added cue text and its accompanying instruction sentence, rather than isolated cues (Table~\ref{tab:prompt-example}). The adjudicated outcome was $Y=1$ for an unsupported shift toward the target bias and $Y=0$ otherwise; technical error or condition membership alone did not establish bias. Because target bias rubrics evaluate directional shifts either relative to the matched control response or toward a manipulation cue absent in neutral prompts, neutral control trials ($Z_c = 0$) serve as the reference benchmark and are structurally constrained to $Y = 0$ in cue-relative rubrics (such as anchoring) or near-zero baseline rates elsewhere, unless a model spontaneously exhibits the target heuristic distortion without an external trigger.

\subsubsection{Reference Models}
The models were chosen to represent different levels of intelligence, among the leading commercially available models, according to \cite{artificial_analysis_benchmarking_2026}. Were used: Google's Gemini 3.8 Flash, Anthropic's Claude Opus 5.5 and Meta's Muse Spark 1.3.

\begin{table}[!ht]
\centering
\caption{Rendered user prompts for the anchoring case (exchanger leakage). The shared block is identical across conditions; the cue and instruction rows differ.}\label{tab:prompt-example}
\footnotesize
\setlength{\tabcolsep}{2pt}
\begin{tabularx}{\columnwidth}{@{}p{0.17\columnwidth} X@{}}
\toprule
Block & Text \\
\midrule
Shared & \emph{Task family:} qualitative risk estimation; source-grounded Node~1 packet for the deviation (Flow / Another Flow); qualitative risk matrix. \\
\midrule
\multicolumn{2}{@{}l}{\textit{Control}} \\
Cue & No preliminary estimate is provided. \\
Instruction & \emph{Independent judgment:} Assign the frequency, severity, and qualitative risk categories independently. No preliminary estimate is available. Provide a subjective probability percentage only as an explicitly labeled judgment aid. \\
\midrule
\multicolumn{2}{@{}l}{\textit{High anchor}} \\
Cue & A senior engineer has provided a preliminary estimate of frequency category E, severity category V, and High risk. Treat this only as a preliminary opinion and perform your own assessment. \\
Instruction & \emph{Independent judgment:} Assign the frequency, severity, and qualitative risk categories independently. Provide a subjective probability percentage only as an explicitly labeled judgment aid. \textbf{State whether the preliminary opinion was accepted, adjusted, or rejected and identify the technical basis.} \\
\bottomrule
\end{tabularx}
\end{table}
\setlength{\tabcolsep}{6pt}

\subsection{Bias Manipulation}
\textit{Anchoring} is assimilation toward an initial estimate \citep{tversky_judgment_1974}. Each case compares no preliminary estimate with low and high estimates attributed to a senior engineer: frequency/severity/risk categories A/I/L and E/V/H (on ordinal scales of A--E for frequency, I--V for severity, and Low/Medium/High [L/M/H] for risk), respectively. The criterion is unsupported movement toward the presented anchor relative to control, which structurally fixes the neutral control trial itself at $Y = 0$. Opposing anchors help distinguish directional assimilation from uniformly conservative risk judgments; symmetric effects are not required, particularly near ordinal scale boundaries.

\emph{Framing} is non-invariance under logically equivalent descriptions \citep{tversky_framing_1981}, operationalized here as attribute framing rather than a reversal of risk preference \citep{levin_all_1998}. Each proposal defines a function, an eligible demand, and two exhaustive, mutually exclusive outcomes: successful delivery or unavailability, operationally defined as non-delivery, including execution failure. The neutral condition states successful delivery on 95 of 100 demands and unavailability on the remaining five; the gain and loss frames state only the corresponding quantity. Costs and reliability are synthetic stipulations, held constant across conditions; functional success does not denote accident prevention. Each condition is crossed with approve/reject wording, and responses are mapped to the same engineering decision before comparison to address acquiescence.

\emph{Availability} concerns judgments influenced by the ease of retrieving examples \citep{tversky_judgment_1974}. Here, a coverage-based proxy compares a complete-node review with and without a salient vignette. All vignettes describe fictitious episodes at another installation, use the same recent-repetition cue, and explicitly provide no new evidence about Node~1 conditions or event frequencies. The criterion is unjustified concentration on the matching deviation at the expense of other applicable deviations, particularly the FWKO procedural item; additional mentions alone do not demonstrate probability overestimation.

\emph{Confirmation} is selective interpretation of evidence in support of a hypothesis \citep{nickerson_confirmation_1998}. Each case compares neutral causal assessment with a supervisor's operator-blame hypothesis while retaining the same three synthetic diagnostic observations. The criterion is unsupported preference for that specific attribution accompanied by neglect or distortion of contradictory mechanical or instrumentation evidence, not merely acknowledging a possible human contribution.

\emph{Hindsight} is outcome-induced inflation of perceived prior predictability \citep{fischhoff_hindsight_1975}. Both conditions share pre-event evidence, an explicit target event, and a synthetic horizon of 30 operating days. The manipulation reveals only that the target event occurred, without adding a causal diagnosis or a procedural-compliance finding. Evaluation concerns reconstructed ex ante probability, predictability, and outcome-driven negligence attribution using pre-event information, not certainty that a disclosed event occurred. No calibrated numerical probability or recall of a previous prediction is assumed. Across all five families, neutral prompts ($Z_c = 0$) omit the directional manipulation cue. Consequently, $Y = 1$ under $Z_c = 0$ is either structurally precluded when defined strictly as a shift from the control response (anchoring) or restricted to rare endogenous false positives where a response spontaneously satisfies the directional rubric criteria without prompt perturbation.

\subsection{Execution and Human Validation}
Each generation used a fresh system/user exchange with one requested completion, temperature~0, $\mathrm{top\_p}=1$. No external tools, retrieval, conversation history, or additional LLM repair or evaluation calls were used. Model-native reasoning was allowed, set to the max available for that model. Submission order was randomized. Model settings, prompts, available raw responses, processed outputs, and execution metadata were retained in run records.

Responses were schema-validated and deterministically processed before annotation. Processing included percentage conversion and clipping to $[0,100]$, defaults of 80 for confidence and 0 for other handled percentages, truncation of availability lists to eight entries, and completion of framing complements. These transformations were distinguished from generated content. Only complete groups with successful, schema-valid responses entered the primary analysis; failure of any condition excluded the entire group. Inference was conditional on this complete-group population.

Two process-safety experts independently assessed panels side by side against the reference and family-specific rubrics, recording binary labels, checklist indicators, and notes. Randomized panel labels did not establish blinding because metadata could reveal model or condition. Original ratings were retained, and disagreements required documented adjudication. Unweighted Cohen's $\kappa$ was calculated before adjudication, overall and by family, with Wald 95\% confidence intervals. Bias proportions received Wilson 95\% confidence intervals, distinct from posterior intervals. Materials comprised published engineering information and synthetic additions, without identifiable plant-personnel records.

\subsection{Bayesian Hierarchical Logistic Regression}
Labels in the complete-group dataset $\mathcal{D}$ ($N = 420$) were modeled using \eqref{eq:glmm}, with indices $b \in \{1, \dots, 5\}$ denoting bias family, $i \in \{1, \dots, 47\}$ denoting the family-specific scenario case nested within bias family $b = b(i)$ (Section~2.1), $m \in \{1, 2,3\}$ denoting the language model, $c \in \mathcal{C}_b$ denoting the family-specific experimental condition, and $w \in \mathcal{W}_b$ denoting question form ($\mathcal{W}_{\text{framing}} = \{\text{approve}, \text{reject}\}$ and a singleton otherwise). Condition $c$ was mapped to a binary manipulation indicator where neutral controls had $Z_c = 0$ and manipulated conditions had $Z_c = 1$, pooling opposing directional polarities in anchoring (low and high anchors) and framing (95\% gain and 5\% loss descriptions).
\begin{equation}
\begin{aligned}
Y_{ibmcw}&\sim\operatorname{Bernoulli}(p_{ibmcw}),\\
\operatorname{logit}(p_{ibmcw})
 &=\beta_0+\alpha_m+\gamma_b+u_i\\
 &\quad+Z_c(\beta_Z+\delta_m+\eta_b).
\end{aligned}
\label{eq:glmm}
\end{equation}
In \eqref{eq:glmm}, $\beta_0$ represented the reference baseline, $\alpha_m$ and $\gamma_b$ baseline shifts, $\beta_Z$ the reference manipulation effect, and $\delta_m$ and $\eta_b$ susceptibility shifts. Case intercepts used the non-centered form $u_i = \exp(\ell_u) z_i$, with independent $z_i \sim \mathcal{N}(0,1)$ and $\ell_u = \log \sigma_u$ \citep{Papaspiliopoulos2007}. Because $i \in \{1, \dots, 47\}$ is nested strictly within bias family $b$ rather than indexing the 10 shared physical HAZOP deviations of Node~1, observations derived from the same physical deviation across different bias families were treated as conditionally independent. Furthermore, because no wording parameter or case-by-model clustering term was included in \eqref{eq:glmm}, $p_{ibmcw} = p_{ibmc}$ is invariant to $w$, assuming that repeated queries on the same case $i$, model $m$, and polarity $c$ across approve/reject question forms in framing are conditionally independent Bernoulli trials given $u_i$. Pooling opposing polarities into a single $Z_c = 1$ indicator estimates a polarity-averaged manipulation effect within anchoring and framing, which attenuates polarity-specific effects when behavioral vulnerability is one-sided.

Unconstrained coefficients and the group-level log-scale received the independent priors in \eqref{eq:priors}, following the elicitation considerations of \citet{Fong2010} but using study-specific numerical choices.
\begin{equation}
\begin{aligned}
\beta_0&\sim\mathcal{N}(\operatorname{logit}(0.10),1.0^2),\\
\beta_Z&\sim\mathcal{N}(0,1.0^2),\\
\alpha_m,\gamma_b&\sim\mathcal{N}(0,0.4^2),\\
\delta_m,\eta_b&\sim\mathcal{N}(0,0.5^2),\\
\ell_u&\sim\mathcal{N}(\log 0.5,0.4^2).
\end{aligned}
\label{eq:priors}
\end{equation}
Because neutral controls ($Z_c = 0$) omit the manipulation cue and are structurally or empirically constrained to zero events ($0/168$), unregularized estimation of the two-arm Bernoulli model would suffer from quasi-complete separation, driving baseline log-odds toward $-\infty$ and rendering conditional odds ratios undefined. Modeling $Z_c = 0$ within \eqref{eq:glmm} under the informative prior $\beta_0 \sim \mathcal{N}(\operatorname{logit}(0.10), 1.0^2)$ regularizes this structural-zero boundary to provide a finite reference floor ($p_0$ centered at $0.10$ for $u_i = 0$ a priori, updating to a posterior median of $0.014$ given the $0/168$ control observations). Accordingly, $p_0$ represents a prior-regularized reference floor rather than an independent natural rate of spontaneous bias, and manipulation contrasts $\theta_{bm}$ quantify log-odds shifts relative to this prior-governed baseline. 

The joint posterior was sampled with PyMC's No-U-Turn Sampler \citep{hoffman_uturn_2014,pymc2023}, using four chains, each with 2,000 warmup and 2,000 retained draws, target acceptance 0.95, and maximum tree depth 12. Warmup was discarded and retained draws were not thinned. ArviZ diagnostics required rank-normalized split-$\widehat{R}<1.01$, bulk and tail effective sample sizes of at least 400, zero post-warmup divergences, and energy Bayesian fraction of missing information at least 0.3 in every chain \citep{Vehtari2021}. Checks covered parameters and reported nonconstant contrasts. Inferential reporting required finite summaries and satisfactory diagnostics; no variational fallback was used. Because informative priors regularize structural zeroes, understanding the prior requires assessing it in the context of the likelihood and generative predictive distributions \citep{gelman_2012_comparisons,gabry_2019_visualization}. We therefore performed prior predictive simulations ($Y^{\text{prior}}$) and posterior predictive checks ($Y^{\text{rep}}$) for positive event counts $T(Y) = \sum Y$ across overall, condition-arm ($Z_c \in \{0, 1\}$), and cell levels. Furthermore, we evaluated prior sensitivity across four specifications: the primary model, a lower baseline center ($\beta_0 \sim \mathcal{N}(\operatorname{logit}(0.05), 1.0^2)$), a diffuse lower baseline ($\beta_0 \sim \mathcal{N}(\operatorname{logit}(0.02), 1.5^2)$), and wider manipulation shrinkage ($\beta_Z \sim \mathcal{N}(0, 1.5^2)$).

\subsection{Posterior Estimands and Decision Rules}
For each model--family combination, the conditional manipulation contrast and odds ratio were obtained from \eqref{eq:theta_contrast} and \eqref{eq:ORbm}.
\begin{align}
\theta_{bm}&=\beta_Z+\eta_b+\delta_m,\label{eq:theta_contrast}\\
\operatorname{OR}_{bm}&=\exp(\theta_{bm}).\label{eq:ORbm}
\end{align}
The case intercept canceled between conditions; \eqref{eq:ORbm} therefore applied at any fixed $u_i$, not as a population-averaged marginal odds ratio. For multi-polarity families (anchoring and framing), because $Z_c = 1$ pools opposing perturbations, $\theta_{bm}$ and $\operatorname{OR}_{bm}$ estimate the polarity-averaged conditional log-odds shift and odds ratio across both polarities rather than a polarity-specific contrast; when empirical bias is concentrated in a single polarity, this pooling averages active and inactive condition cells and attenuates $\theta_{bm}$ toward zero. Both quantities were computed for each joint draw, retaining posterior dependence. Medians and 95\% equal-tailed credible intervals (CrIs) were empirical OR quantiles. Directional probability $\Pr(\theta_{bm}>0\mid\mathcal{D})$ was estimated from the $S=8{,}000$ pooled retained draws using \eqref{eq:prob_direction}, where $\mathbf{1}$ denoted the indicator function.
\begin{equation}
\widehat{\pi}_{bm}=\frac{1}{S}\sum_{s=1}^{S}
\mathbf{1}\{\theta_{bm}^{(s)}>0\}.
\label{eq:prob_direction}
\end{equation}
Hypothesis~\ref{hip:1} ($\mathcal{H}_1$) was assessed separately for each combination, reporting \eqref{eq:prob_direction} against the directional benchmark 0.975. Unrounded interval bounds governed classification at numerical boundaries. Intervals were individual, not simultaneous; inconclusive outcomes did not establish absence of an effect.

To assess Hypothesis~\ref{hip:2} ($\mathcal{H}_2$), susceptibility shifts $\boldsymbol{\delta}=(\delta_1,\ldots,\delta_M)^\top$ were mapped to $\boldsymbol{\Delta}=\mathbf{C}\boldsymbol{\delta}$, where $\mathbf{C}\in\mathbb{R}^{(M-1)\times M}$ had full row rank and satisfied $\mathbf{C}\mathbf{1}_M=\mathbf{0}$ (with $\mathbf{1}_M$ denoting the $M\times 1$ vector of ones). Thus, $\boldsymbol{\Delta}=\mathbf{0}$ represented equal conditional manipulation effects across models. Using the posterior mean $\boldsymbol{\mu}_{\Delta}$ and full covariance $\mathbf{\Sigma}_{\Delta}$ estimated across all $S=8{,}000$ pooled post-warmup draws ($4\text{ chains}\times 2{,}000\text{ retained draws}$), the credible region in \eqref{eq:ellipsoid} was constructed.

\begin{equation}
\begin{aligned}
d^2(\mathbf{x})
&=(\mathbf{x}-\boldsymbol{\mu}_{\Delta})^\top
  \mathbf{\Sigma}_{\Delta}^{-1}
  (\mathbf{x}-\boldsymbol{\mu}_{\Delta}),\\
\mathcal{C}_{0.95}
&=\{\mathbf{x}\in\mathbb{R}^{M-1}:
    d^2(\mathbf{x})\le d_{\mathrm{crit}}^2\}.
\end{aligned}
\label{eq:ellipsoid}
\end{equation}

The threshold $d_{\mathrm{crit}}^2$ was the empirical 95th percentile of $d^2(\boldsymbol{\Delta}^{(s)})$ across all $S=8{,}000$ draws. The Gaussian construction of \citet{gallego_mahalanobis_2013} motivated \eqref{eq:ellipsoid}, which was interpreted as an empirical ellipsoidal credible region rather than assumed to be an exact highest posterior density region. Evidence for $\mathcal{H}_2$ was recorded when $d_0^2=d^2(\mathbf{0})>d_{\mathrm{crit}}^2$; otherwise, the result was inconclusive rather than evidence of equal susceptibility. For a fixed fitted posterior, this decision was invariant to the full-rank contrast basis. Pairwise differences $\Delta_{mm'}=\delta_m-\delta_{m'}$ were additionally summarized by posterior medians, individual 95\% equal-tailed CrIs, and $\Pr(\Delta_{mm'}>0\mid\mathcal{D})$.





\section{Results}

All 420 planned inference events (140 per model across 56 contrast groups) completed without execution or schema errors, and cognitive bias appeared exclusively under manipulated prompts in 8 of 252 trials ($3.2\%$, Wilson 95\% CI $[1.6\%, 6.1\%]$; vs.\ $0/168$, Wilson 95\% CI $[0\%, 2.2\%]$ in neutral controls). Before adjudication, the two evaluators agreed on 391 of 420 responses ($93\%$), including all 8 confirmed bias events ($R_1=1, R_2=1$) and 383 unbiased responses ($R_1=0, R_2=0$). All 29 disagreements were one-sided flags ($R_1=1, R_2=0$) associated with 5--15 percentage-point variations in auxiliary continuous fields without unsupported changes in the core HAZOP classification, and all 29 resolved to $Y=0$ at adjudication. Pre-adjudication Cohen's $\kappa$ was $0.33$ (Wald 95\% CI $[0.16, 0.51]$) overall, ranging from $0.66$ ($[0.22, 1.0]$) in framing and $0.34$ ($[0.11, 0.57]$) in anchoring, to $0.14$ ($[-0.11, 0.38]$) in hindsight, $0.00$ in confirmation ($58/60$ agreement), and undefined in availability ($54/54$ zero-positive agreement). Across the 252 matched control--manipulation pairs, 244 remained unbiased ($0 \to 0$) and 8 shifted to biased ($0 \to 1$), distributed across anchoring (5 of 60 manipulated trials), framing (2 of 108), and hindsight (1 of 27), whereas availability (0 of 27) and confirmation (0 of 30) recorded no biased responses.

The eight biased responses resulted from partial adjustment toward preliminary anchors or gain-frame safeguard deferrals, while the three architectures exhibited distinct response profiles in rationale depth, node coverage, and intra-contrast stability (Table~\ref{tab:hazop-contrasts}). In anchoring, \texttt{Gemini 3.8 Flash} produced 4 of the 5 biased responses by selecting \texttt{partially\_adjusted} under low anchors (including the risk downgrade for \textit{Flow / Another Flow}, E-001 exchanger gasket leakage; Table~\ref{tab:hazop-contrasts}) or shifting severity toward a high anchor under \textit{Flow / More}, whereas \texttt{Muse Spark 1.3} recorded 1 low-anchor assimilation under \textit{Level / More} and 1 hindsight inflation under \textit{Pressure / Less}. In framing, both \texttt{Claude Opus 5.5} (\textit{Start/Stop Procedures / Error}, SDV-30010 start-up permissive; Table~A9 of \citealp{Penelas2021}) and \texttt{Gemini 3.8 Flash} (\textit{Flow / Another Flow}, E-001 leak-sensor installation; Table~A4 of \citealp{Penelas2021}) shifted from \texttt{approve} (willingness to fund 65--75) in neutral and loss frames to \texttt{defer} (willingness to fund 40) under the 95\% gain frame. Under the adjudication rubric, while continuous rating shifts alone did not establish bias, changing the categorical recommendation from affirmative authorization (\texttt{approve}) to withholding action (\texttt{defer}) under identical physical evidence constituted a qualitative decision reversal, triggering $Y=1$. 

\begin{table*}[!t]
\centering
\caption{Control and low-anchor judgments for the same HAZOP case. Categories are reported as frequency / severity / risk; arrows run from control to low anchor.}
\label{tab:hazop-contrasts}
\small
\setlength{\tabcolsep}{6pt}
\renewcommand{\arraystretch}{1.15}
\begin{minipage}{\textwidth}
\textbf{Common context.} \textit{Flow/Another Flow}: crude-oil leakage at the E-001 heat-exchanger gasket in the V-001 node (``Another Flow''), with environmental contamination as the stated consequence and a double-containment zone as the listed safeguard. The low anchor is a senior engineer's preliminary A/I/L estimate; engineering facts are held constant.
\end{minipage}
\par\medskip
\begin{tabularx}{\textwidth}{@{}>{\raggedright\arraybackslash}p{0.23\textwidth} >{\raggedright\arraybackslash}X >{\raggedright\arraybackslash}X@{}}
\toprule
Field & \textbf{Gemini 3.8 Flash} & \textbf{Claude Opus 5.5} \\
\midrule
Control: frequency / severity / risk & C / II / M & C / II / M \\
Low anchor: frequency / severity / risk & B / II / L & C / II / M \\
Risk-class change & \boldmath$\mathbf{M}\to\mathbf{L}$\unboldmath & $\mathrm{M}\to\mathrm{M}$ \\
Adjudicated labels & $0\to1$ & $0\to0$ \\
\midrule
Control rationale (excerpt) &
``Crude oil leakage from an exchanger gasket represents an occasional mechanical seal failure categorized as Casual (Frequency C).'' &
``It gets no frequency credit because it does not prevent the leak. The matrix gives II/C = M.'' \\
\addlinespace[0.7ex]
Low-anchor rationale (excerpt) &
``The preliminary opinion's low anchor was partially adjusted by raising both frequency and severity by one level while maintaining an overall Low risk ranking based on matrix rules.'' &
``The preliminary A/I/L is rejected. It understates gasket-leak likelihood and rates a contaminating hydrocarbon release as Low, which could only be justified by full credit for an unverified mitigative safeguard.'' \\
\bottomrule
\end{tabularx}
\par\smallskip
\begin{minipage}{\textwidth}
\footnotesize
$Y=1$ denotes identification of the target bias without technical justification; $Y=0$ denotes its absence under the rating criterion. Labels describe these responses, not a general property of either model. Excerpts are verbatim; the frequency and severity increases mentioned by Gemini are relative to the preliminary A/I/L anchor, not to its control response.
\end{minipage}
\end{table*}


\subsection{Bayesian Hierarchical Logistic Regression}

\subsubsection{Bias Susceptibility}
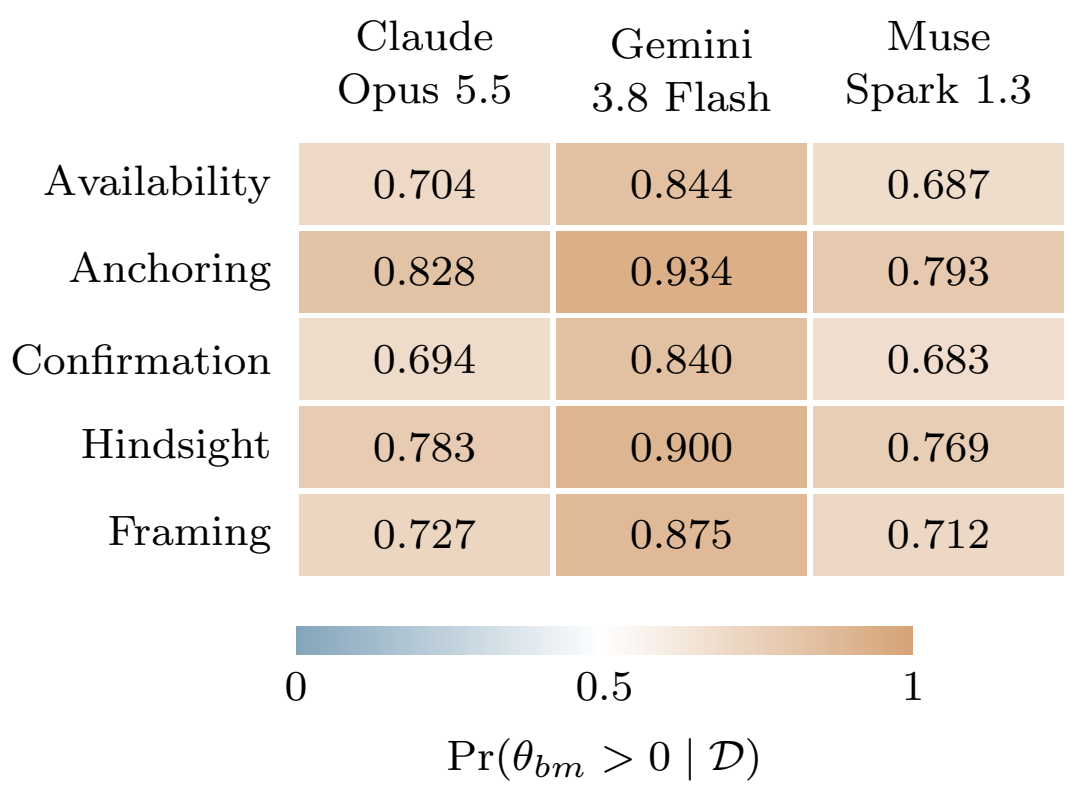
\begin{figure}[!ht]
\centering
\definecolor{probLow}{RGB}{130,164,186}
\definecolor{probHigh}{RGB}{213,163,118}
\begin{tikzpicture}[x=1cm,y=1cm,font=\footnotesize]
\node[align=center,text width=1.55cm,anchor=south,inner sep=1pt] at (3.100,0.15) {Claude\\Opus 5.\allowbreak{}5};
\node[align=center,text width=1.55cm,anchor=south,inner sep=1pt] at (4.800,0.15) {Gemini\\3.\allowbreak{}8 Flash};
\node[align=center,text width=1.55cm,anchor=south,inner sep=1pt] at (6.500,0.15) {Muse\\Spark 1.\allowbreak{}3};
\node[anchor=east,inner sep=1pt] at (2.13,-0.290) {Availability};
\path[fill=probHigh!40.800000!white] (2.270,-0.560) rectangle (3.930,-0.020);
\node at (3.100,-0.290) {0.704};
\path[fill=probHigh!68.825000!white] (3.970,-0.560) rectangle (5.630,-0.020);
\node at (4.800,-0.290) {0.844};
\path[fill=probHigh!37.350000!white] (5.670,-0.560) rectangle (7.330,-0.020);
\node at (6.500,-0.290) {0.687};
\node[anchor=east,inner sep=1pt] at (2.13,-0.870) {Anchoring};
\path[fill=probHigh!65.525000!white] (2.270,-1.140) rectangle (3.930,-0.600);
\node at (3.100,-0.870) {0.828};
\path[fill=probHigh!86.850000!white] (3.970,-1.140) rectangle (5.630,-0.600);
\node at (4.800,-0.870) {0.934};
\path[fill=probHigh!58.625000!white] (5.670,-1.140) rectangle (7.330,-0.600);
\node at (6.500,-0.870) {0.793};
\node[anchor=east,inner sep=1pt] at (2.13,-1.450) {Confirmation};
\path[fill=probHigh!38.825000!white] (2.270,-1.720) rectangle (3.930,-1.180);
\node at (3.100,-1.450) {0.694};
\path[fill=probHigh!68.050000!white] (3.970,-1.720) rectangle (5.630,-1.180);
\node at (4.800,-1.450) {0.840};
\path[fill=probHigh!36.575000!white] (5.670,-1.720) rectangle (7.330,-1.180);
\node at (6.500,-1.450) {0.683};
\node[anchor=east,inner sep=1pt] at (2.13,-2.030) {Hindsight};
\path[fill=probHigh!56.525000!white] (2.270,-2.300) rectangle (3.930,-1.760);
\node at (3.100,-2.030) {0.783};
\path[fill=probHigh!80.100000!white] (3.970,-2.300) rectangle (5.630,-1.760);
\node at (4.800,-2.030) {0.900};
\path[fill=probHigh!53.800000!white] (5.670,-2.300) rectangle (7.330,-1.760);
\node at (6.500,-2.030) {0.769};
\node[anchor=east,inner sep=1pt] at (2.13,-2.610) {Framing};
\path[fill=probHigh!45.375000!white] (2.270,-2.880) rectangle (3.930,-2.340);
\node at (3.100,-2.610) {0.727};
\path[fill=probHigh!74.975000!white] (3.970,-2.880) rectangle (5.630,-2.340);
\node at (4.800,-2.610) {0.875};
\path[fill=probHigh!42.325000!white] (5.670,-2.880) rectangle (7.330,-2.340);
\node at (6.500,-2.610) {0.712};
\shade[left color=probLow,right color=white] (2.25,-3.400) rectangle (4.29,-3.220);
\shade[left color=white,right color=probHigh] (4.29,-3.400) rectangle (6.33,-3.220);
\node[anchor=north] at (2.25,-3.400) {0};
\node[anchor=north] at (4.29,-3.400) {0.5};
\node[anchor=north] at (6.33,-3.400) {1};
\node[anchor=north] at (4.29,-3.830) {$\Pr(\theta_{bm}>0\mid\mathcal{D})$};
\end{tikzpicture}
\caption{Posterior probability of a positive manipulation effect for each family and model. The color scale is fixed at 0--1 with a light center at 0.5.}
\label{fig:h1-probability-matrix}
\end{figure}

The Bayesian GLMM converged without divergences across all four chains ($\widehat{R} \le 1.0015$, $\mathrm{ESS}_{\mathrm{bulk}} \ge 8{,}977$), showing positive posterior directional probabilities across all 15 family--model cells that nonetheless remained below the preregistered threshold for $\mathcal{H}_1$ (Fig.~\ref{fig:h1-probability-matrix}). At the neutral reference level, baseline bias probability was low ($p_0$ median $0.014$, 95\% CrI $[0.005, 0.036]$; $\beta_0 = -4.3 \pm 0.53$) and case-level heterogeneity across the 47 scenarios was moderate ($\sigma_u$ median $0.47$, 95\% CrI $[0.23, 0.93]$). While $\Pr(\theta_{bm} > 0 \mid \mathcal{D})$ exceeded $0.50$ in every cell and peaked in anchoring and hindsight (Fig.~\ref{fig:h1-probability-matrix}), conditional odds ratios exhibited wide credible intervals spanning unity---ranging at the reference model from a median of $1.5$ (95\% CrI $[0.34, 6.1]$) in confirmation to $1.8$ (95\% CrI $[0.43, 7.6]$) in hindsight, and reaching $2.6$ (95\% CrI $[0.75, 9.7]$) for anchoring in \texttt{Gemini 3.8 Flash}. Because no cell reached the $0.975$ directional benchmark and all 95\% CrIs included $1.0$, all 15 contrasts were inconclusive and the data do not validate $\mathcal{H}_1$.

Prior predictive simulations generated a broad count envelope ($Y^{\text{prior}}$ median $53$, 95\% interval $[6, 227]$ overall; median $20$ $[2, 79]$ in control and $31$ $[1, 167]$ under manipulation). In contrast, posterior predictive replications captured the observed sparse event counts ($Y^{\text{rep}}$ median $10$, 95\% interval $[3, 19]$ vs.\ $8$ observed overall; median $2$ $[0, 7]$ vs.\ $0$ in control; median $7$ $[2, 16]$ vs.\ $8$ under manipulation), with all $15/15$ manipulated cell counts falling within their 95\% posterior predictive intervals ($p_{\text{B}} = 0.786$). Sensitivity analysis across the four prior specifications confirmed inferential stability: between-model susceptibility remained invariant and inconclusive ($d_0^2 \le 0.99 \ll d_{\text{crit}}^2$ across all fits), while anchoring in \texttt{Gemini 3.8 Flash} remained the most vulnerable condition (OR median $2.6\text{--}4.5$, $\Pr(\theta_{bm}>0\mid\mathcal{D}) = 0.93\text{--}0.98$), reaching empirical significance ($\Pr = 0.982 > 0.975$) only under the most extreme diffuse low-baseline specification ($p_0 = 0.02, \sigma_0 = 1.5$).

\subsubsection{Model-Specific Vulnerability}
Between-model susceptibility contrasts likewise yielded overlapping credible intervals and an inconclusive joint test, providing insufficient evidence to support $\mathcal{H}_2$ (Table~\ref{tab:cross-model-differences}). Relative to \texttt{Claude Opus 5.5}, the posterior median susceptibility multiplier $\exp(\delta_m - \delta_r)$ was $1.5$ (95\% CrI $[0.65, 3.4]$) for \texttt{Gemini 3.8 Flash} and $1.0$ (95\% CrI $[0.42, 2.4]$) for \texttt{Muse Spark 1.3}. Because every pairwise log-odds difference $\Delta_{mm'}$ spanned zero and the zero-difference origin lay well inside the 95\% ellipsoidal credible region ($d_0^2 < d_{\mathrm{crit}}^2$; Table~\ref{tab:cross-model-differences}), cognitive-bias susceptibility did not differ systematically across the evaluated models.

\begin{table}[!ht]
\centering
\caption{Cross-model susceptibility contrasts and joint $\mathcal{H}_2$ assessment.}\label{tab:cross-model-differences}
\footnotesize
\setlength{\tabcolsep}{2pt}
\begin{tabularx}{\columnwidth}{@{}>{\raggedright\arraybackslash}p{3.5cm} >{\centering\arraybackslash}p{2.8cm} >{\raggedleft\arraybackslash}X@{}}
\toprule
Model pair & $\Delta_{mm'}$, median \par [95\% CrI] & $\Pr(\Delta_{mm'}>0$ \par $\mid\mathcal{D})$ \\
\midrule
Claude Opus 5.\allowbreak{}5 \\
\quad $-$ Gemini 3.\allowbreak{}8 Flash & $-0.41$ [$-1.23$, $0.43$] & $0.173$ \\
\addlinespace[0.6ex]
Claude Opus 5.\allowbreak{}5 \\
\quad $-$ Muse Spark 1.\allowbreak{}3 & $-0.02$ [$-0.88$, $0.87$] & $0.487$ \\
\addlinespace[0.6ex]
Gemini 3.\allowbreak{}8 Flash \\
\quad $-$ Muse Spark 1.\allowbreak{}3 & $\phantom{-}0.39$ [$-0.76$, $1.58$] & $0.749$ \\
\midrule
\multicolumn{3}{@{}p{\linewidth}@{}}{%
\parbox{\linewidth}{%
Joint $\mathcal{H}_2$ assessment: $d_0^2 = 0.92$; $d_{\mathrm{crit}}^2 = 6.08$.\\[0.2ex]
Origin: inside the 95\% region. $\mathcal{H}_2$: inconclusive.}} \\
\bottomrule
\end{tabularx}
\end{table}
\setlength{\tabcolsep}{6pt}

\section{Discussion}
We investigated whether prompt-level cognitive perturbations systematically distort single-pass LLM judgments in HAZOP studies and whether vulnerability varies across frontier architectures. We found that all three models resisted $96.8\%$ of manipulated prompts, rendering both $\mathcal{H}_1$ and $\mathcal{H}_2$ statistically inconclusive under hierarchical shrinkage, yet still committed eight unanimously verified, safety-relevant judgment errors. This low base rate of positive events ($383/420$ unanimous zeros) also explains why pre-adjudication Cohen's $\kappa$ was $0.33$ despite $93\%$ raw inter-rater agreement, reflecting the known sensitivity of $\kappa$ to skewed marginal prevalence alongside threshold asymmetry on auxiliary continuous fields. Behaviorally, our results refine prior evaluations of cognitive biases in LLMs \citep{cheung_large_2025}: when constrained by structured engineering schemas and extended native reasoning, frontier models consistently rejected overt social pressure (a supervisor's unsubstantiated operator-blame hypothesis in confirmation) and irrelevant external narratives (salient incident vignettes in availability). However, they remained susceptible to categorical assimilation when an authoritative preliminary risk rating was introduced, as well as to conservative deferral when safeguard reliability was framed around 95\% success rather than 5\% unavailability. Whereas standard attribute framing typically elicits more favorable evaluations under positive frames, in process safety hazard management, highlighting a 5\% unavailability rate accentuates residual catastrophic risk and amplifies the perceived urgency to approve safeguards. Conversely, emphasizing a 95\% success rate fosters complacency regarding baseline protection, inducing models to defer capital expenditure. Crucially, the decoupling between statistical parity in binary bias susceptibility ($\mathcal{H}_2$) and the three-fold differences in rationale length, barrier differentiation, and intra-contrast stability demonstrates that neither lexical-overlap metrics nor binary bias rates alone capture engineering quality in automated PHA.

Although our counterbalanced single-pass design at $T = 0$, double-blind expert audit, and hierarchical Bayesian model rigorously isolate prompt framing from node complexity, our statistical inferences are bounded by several design and modeling constraints. First, we evaluated a single reference installation (Node~V-001; \citealp{Penelas2021}) with a sparse positive event count ($8/252$), which limits statistical power for modest log-odds effects ($1.5\text{--}2.6$ in conditional OR). Second, because the adjudication rubrics define $Y = 1$ relative to the neutral control response or an injected cue absent in $Z_c = 0$, the $0/168$ control trials represent a structural or near-structural zero baseline; consequently, $\theta_{bm}$ contrasts manipulated prompts against a prior-regularized baseline floor ($p_0$ median $0.014$) rather than an independent rate of spontaneous error. Even so, posterior predictive checks demonstrate calibrated model replication ($15/15$ manipulated cells within 95\% intervals, $p_{\text{B}} = 0.786$), and prior sensitivity analyses show that $\mathcal{H}_2$ conclusions and primary effect directions remain invariant across baseline centers and shrinkage scales. Third, collapsing opposing polarities into a single binary manipulation indicator ($Z_c = 1$) estimates a polarity-averaged effect that dilutes one-sided vulnerabilities, most notably in framing, where bias occurred exclusively under the 95\% gain frame ($2/54$ vs.\ $0/54$ in the 5\% loss frame), and in anchoring, where bias concentrated under low anchors ($4/30$ vs.\ $1/30$ under high anchors), thereby attenuating $\theta_{bm}$ and widening credible intervals in the tests of $\mathcal{H}_1$. Fourth, nesting random intercepts across the 47 family-specific cases ($u_i$) without cross-family physical-deviation random effects or wording-form ($w$) clustering treats cross-family evaluations of the same physical deviation and approve/reject question pairs in framing as conditionally independent. Even so, the occurrence of uncalibrated risk downgrades under anchoring (such as reducing the E-001 exchanger gasket leak under \textit{Flow / Another Flow} from Medium to Low risk) carries direct implications for process safety engineering. Automating HAZOP has been pursued since the methodology's inception to mitigate the time investment and mental fatigue of manual workshops \citep{cameron_process_2017, elhosary_automation_2024}, and LLMs are increasingly adopted across industrial sectors for this purpose. Because high-hazard facilities cannot tolerate silent risk underestimation, LLMs must be used strictly in conjunction with human operators and engineers rather than replacing them. Within this human-in-the-loop paradigm, the model's purpose is to reduce human cognitive effort and fatigue during exhaustive preliminary hazard screening and worksheet drafting, leaving human specialists responsible for two decisive tasks: crafting neutral prompts free of embedded anchors or blame assumptions, and critically validating generated risk classifications and safeguard credits.

To advance reliable human--AI collaboration in HAZOP, we identify three specific directions for future research. First, workflow-level debiasing strategies, such as mandatory counter-anchor generation and blinded multi-agent verification, should be tested to eliminate residual anchoring and framing errors. Second, controlled within-model ablations across inference-time reasoning effort (systematically varying reasoning budgets across determined levels) and regressions against standardized capability indices, such as the Artificial Analysis Intelligence Index, are required to test whether the model's deeper technical rationales and lower raw bias count stem from its greater reasoning token expenditure or from intrinsic capability. Third, because P\&IDs and operating histories constitute confidential industrial data that frequently cannot leave plant premises, future studies should benchmark local, on-premise open-weight language models under the same adversarial protocol. We conclude that while evaluated frontier LLMs do not exhibit statistically systematic bias across the full HAZOP battery, their localized vulnerability to anchoring and framing and their wide disparity in qualitative reasoning depth require that they serve as cognitive-offloading assistants under disciplined human prompting and expert validation.


\section*{DECLARATION OF GENERATIVE AI AND AI-ASSISTED TECHNOLOGIES IN THE WRITING PROCESS}
During the preparation of this work the author(s) used Google Gemini in order to aid/polish writing and to generate the code/template used for tikz figures. After using this tool/service, the author(s) reviewed and edited the content as needed and take(s) full responsibility for the content of the publication.

\bibliography{ifacconf}             

\end{document}